\documentclass[conference,a4paper]{IEEEtran}

\usepackage{amsmath,epsfig}

\newcommand{\bel}{\mathrm{bel}}
\newcommand{\pl}{\mathrm{pl}}
\newcommand{\betP}{\mathrm{betP}}

\begin{document}
%
\title{Experts Fusion and Multilayer Perceptron Based on Belief Learning for Sonar Image Classification}

\author{Arnaud Martin\\
ENSIETA\\
$\mbox{E}^3\mbox{I}^2$ EA3876\\
2 rue Fran{\c c}ois Verny, 29806 Brest Cedex 09, France\\
Email: Arnaud.Martin@ensieta.fr
\and
Christophe Osswald\\
ENSIETA\\
$\mbox{E}^3\mbox{I}^2$ EA3876\\
2 rue Fran{\c c}ois Verny, 29806 Brest Cedex 09, France\\
Email: Christophe.Osswald@ensieta.fr}

\maketitle

\begin{abstract}
The sonar images provide a rapid view of the seabed in order to characterize it. However, in such as uncertain environment, real seabed is unknown and the only information we can obtain, is the interpretation of different human experts, sometimes in conflict. In this paper, we propose to manage this conflict in order to provide a robust reality for the learning step of classification algorithms. The classification is conducted by a multilayer perceptron, taking into account the uncertainty of the reality in the learning stage. The results of this seabed characterization are presented on real sonar images.

\end{abstract}

\IEEEpeerreviewmaketitle

\section{Introduction}

\begin{figure}[htb]
\begin{center}
\includegraphics[height=4.7cm]{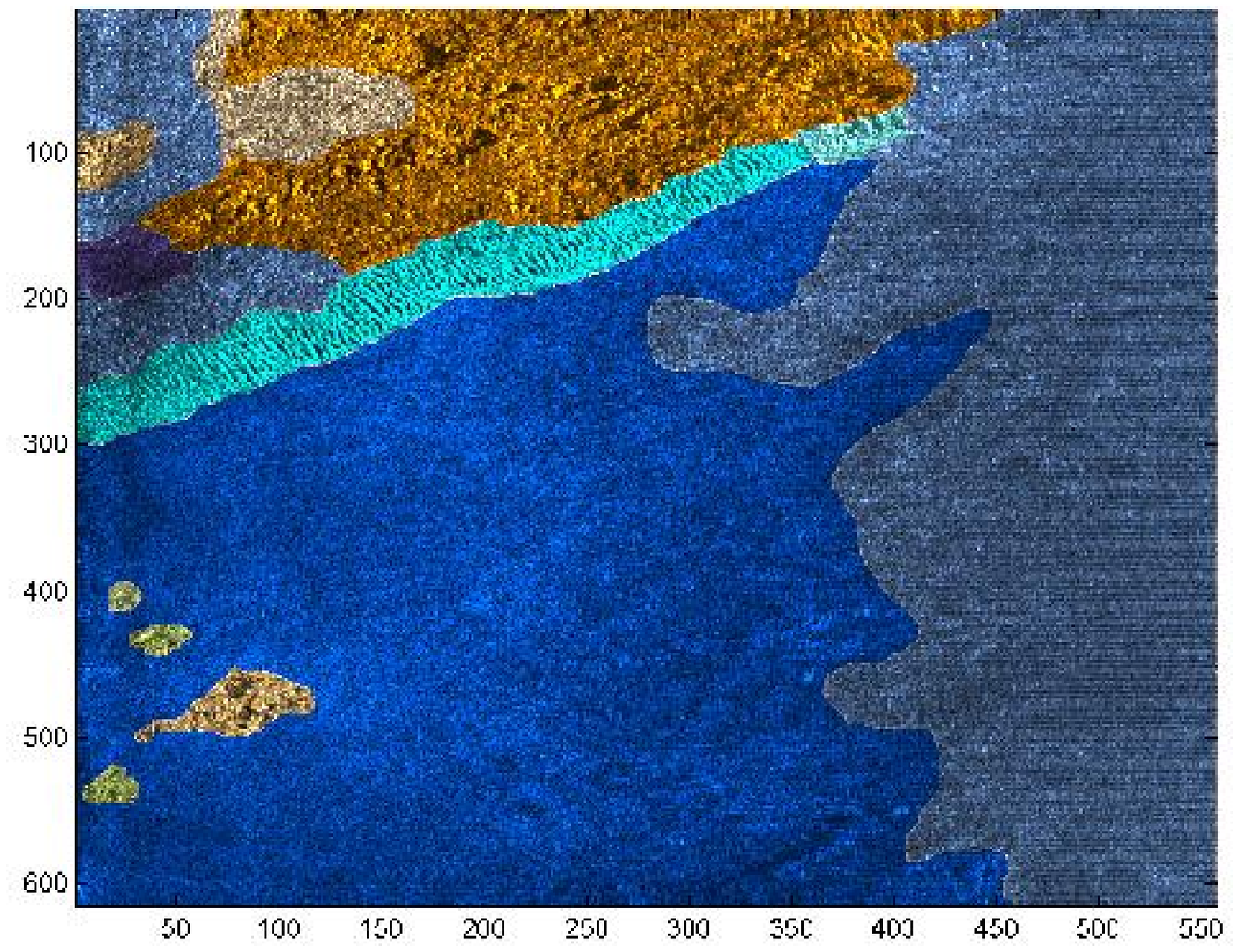}
\includegraphics[height=4.7cm]{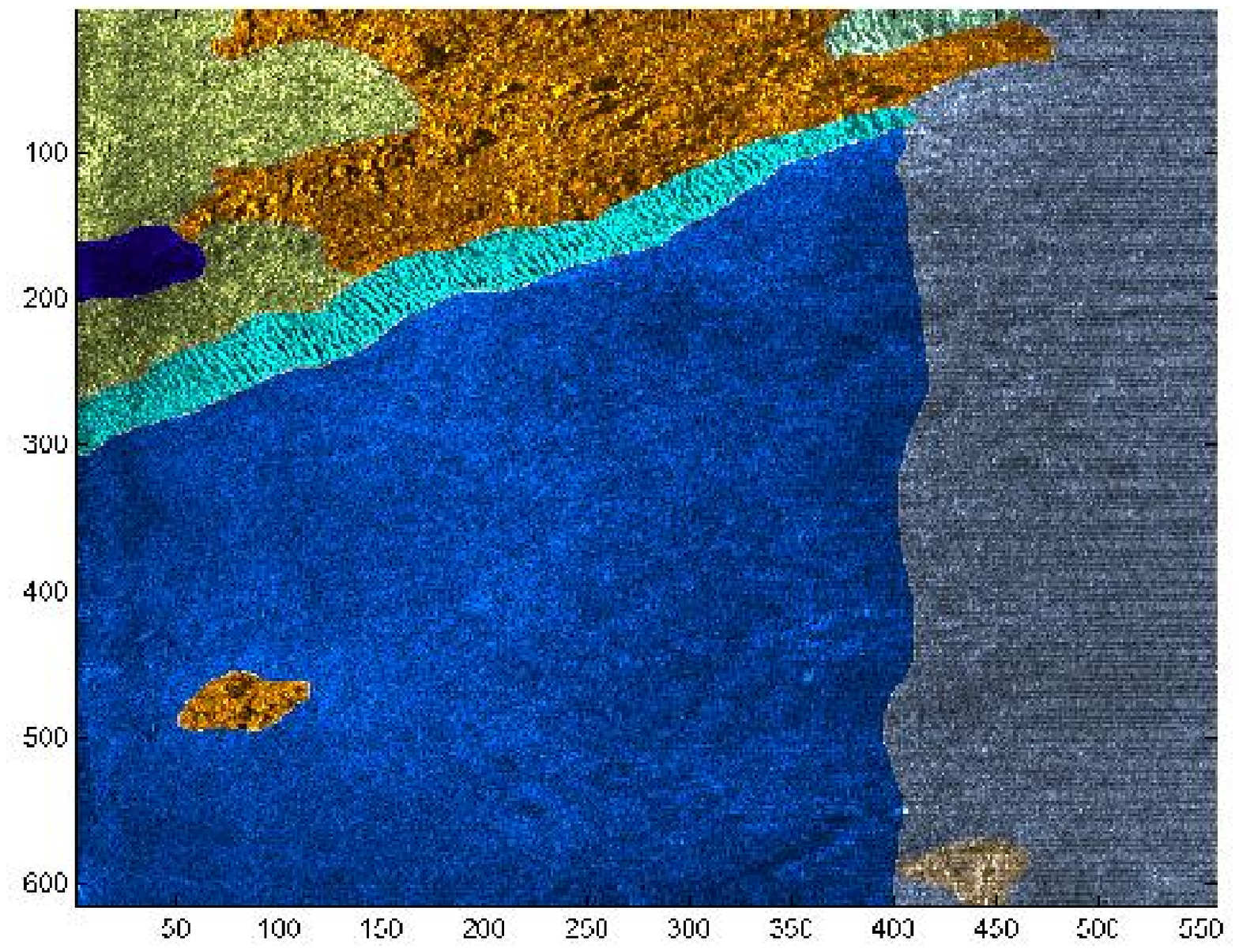}
\end{center}
\vspace{-0.5cm}
\caption{Segmentation given by two experts.}
\label{expert}
\end{figure}

The seabed characterization serves many useful purposes, {\it e.g} help the navigation of Autonomous Underwater Vehicles or provide data to sedimentologists. In such sonar applications, seabed images are obtained with many imperfections \cite{Martin05}. Indeed, in order to build images, a huge number of physical data (geometry of the device, coordinates of the ship, movements of the sonar, etc.) are taken into account, but these data are polluted with a large amount of noises caused by instrumentation. In addition, there are some interferences due to the signal traveling on multiple paths (reflection on the bottom or surface), due to speckle, and due to fauna and flora. Therefore, sonar images have a lot of imperfections such as imprecision and uncertainty; thus sediment classification on sonar images is a difficult problem even for human experts. In this kind of applications, the reality is unknown and different experts can propose different classifications of the image. Figure \ref{expert} exhibits the differences between the interpretation and the certainty of two sonar experts trying to differentiate the type of sediment (rock, cobbles, sand, ripple, silt) or shadow when the information is invisible. Each color corresponds to a kind of sediment and the associated certainty of the expert for this sediment expressed in term of sure, moderately sure and not sure. Thus, in order to learn an automatic classification algorithm, we must take into account this difference and the uncertainty of each expert. For example, how a tile of rock labeled as {\it not sure} must be taken into account in the learning step of the classifier and how to take into account this tile if another expert says that it is sand?

Textured image classification, such as sonar image, is generally done on a local part of the image (pixel, or most of the time on small tiles of {\it e.g.} 16$\times$16 or 32$\times$32 pixels). Usual sonar image classification methods are usually supervised \cite{LeChenadec05,Lianantonakis05, Martin05} and can be described into three steps. First, significant features are extracted from these tiles. Generally, a second step in necessary in order to reduce these features, because they are too numerous. In the third step, these features feed classification algorithms. The particularity in considering small tiles in image classification is that sometimes, two or more classes can co-exist on a tile. How to take into account the tiles with more than one sediment?
    
Many fusion theories can be used for the experts fusion in image classification such as voting rules \cite{Xu92,Lam97}, possibility theory \cite{Zadeh78,Dubois88a}, belief function theory \cite{Dempster67,Shafer76,Smets90,Smets94}. In our case, experts can express their certitude on their perception. As a result, probabilities theories such as the Bayesian theory or the belief function theory are more adapted. Indeed, the possibility theory is more adapted to modelize the imprecise data whereas probability-based theories is more adapted to modelize the uncertain data. Of course both possibility and probability-based theories can imitate imprecise and uncertain data at the same time, but not so easily. That is why our choice is conducted on the belief function theory, also called the Dempster-Shafer theory \cite{Dempster67, Shafer76} or the Transferable Belief Model \cite{Smets90,Smets94}. We can divide the fusion approach into two levels: the credal level and the decision level. The credal level can be described into three stages: the belief function model, the estimation of some parameters depending on the model (not always necessary), and the combination. The most difficult step is presumably the first one: the belief function model from which the other steps follow. 

The paper is organized as follow: in a first section we recall the bases of the transferable belief model. Next, we present an approach of experts fusion in order to obtain a reality on our sonar images. We propose a new multialyer perceptron based on belief learning. In the last section, we show the result of the classification of sonar images.

\section{Transferable Belief Model bases}
\label{TBM}
\subsection{Credal level}

\subsubsection{Belief Function Models}
Consider the space of discernment $\Theta=\{C_1, C_2, \ldots, C_N \}$, where $C_i$ is the hypothesis ``the considered tile belongs to the class $i$''.
The belief functions can be expressed in several forms: the basic belief assignments (bba) $m$, the credibility function
$\bel$ and the plausibility function $\pl$, which are in one-to-one correspondence.

The basic belief assignments (bba) $m$ are defined by the mapping of the power set $2^\Theta$ (defined by all the disjunctions of $\Theta$) onto $[0,1]$, with:
\begin{equation}
\label{normDST}
\sum_{X\in 2^\Theta} m(X)=1.
\end{equation}

In the open world case \cite{Smets90}:
\begin{equation}
\label{open}
m(\emptyset)>0.
\end{equation}

These simple conditions in equation (\ref{open}) and (\ref{normDST}), give a large panel of definitions of the bba, which is one of the difficulties of the theory. The belief functions must therefore be chosen according to the intended application.

The credibility function is given for all $X \in 2^\Theta$ by:
\begin{eqnarray}
\bel(X)=\sum_{Y \in 2^X, Y \neq \emptyset} m(Y).
\end{eqnarray}
The plausibility function is given for all $X \in 2^\Theta$ by:
\begin{eqnarray}
\pl(X)=\sum_{Y \in 2^\Theta, Y\cap X \neq \emptyset} m(Y)=\bel(\Theta)-\bel(X^c),
\end{eqnarray}
where $X^c$ is the complementary of $X$. 

\subsubsection{Combination rules}
Many combination rules have been proposed these last years in the context of the belief function theory (\cite{Yager87, Dubois88, Smets90, Smets93, Martin06}, {\it etc.}). In the context of the TBM, the combination rule most used today seems to be the conjunctive rule given by \cite{Smets90} for all $X \in 2^\Theta$ by:
\begin{eqnarray}
\label{conjunctive}
m_c(X)=\displaystyle \sum_{Y_1 \cap ... \cap Y_M = X} \prod_{j=1}^M m_j(Y_j),
\end{eqnarray}
where $Y_j \in 2^\Theta$ is the response of the expert $j$, and $m_j(Y_j)$ the associated belief function.

However, the conflict (that is given by $m_c(\emptyset)$) can be redistributed on partial ignorance like in the Dubois and Prade rule \cite{Dubois88}, a mixed conjunctive and disjunctive rule given for all $X \in 2^\Theta$, $X\neq \emptyset$ by: 
\begin{eqnarray}
\label{DP}
\begin{array}{l}
m_{DP}(X)=\displaystyle \sum_{Y_1 \cap ... \cap Y_M = X} \prod_{j=1}^M m_j(Y_j)\\

+\displaystyle \sum_{
\begin{array}{c}
\scriptstyle Y_1 \cup ... \cup Y_M = X\\
\scriptstyle Y_1 \cap ... \cap Y_M = \emptyset \\
\end{array}} \prod_{j=1}^M m_j(Y_j),
\end{array}
\end{eqnarray}
where $Y_j \in 2^\Theta$ is the response of the expert $j$, and $m_j(Y_j)$ the associated belief function.

We have proposed another proportional conflict redistribution rule \cite{Martin06} for $M$ experts, for $X\in 2^\Theta$, $X\neq \emptyset$:
\begin{eqnarray}
\label{GeneDSmTcombination}
\begin{array}{l}
m_{PCR}(X)=m_c(X)+\\
\displaystyle
\sum_{i=1}^M m_i(X)^2 . \sum_{\begin{array}{c}
\scriptstyle (Y_{\sigma_i(1)},...,Y_{\sigma_i(M-1)})\in (2^\Theta)^{M-1}\\
\scriptstyle {\displaystyle \mathop{\cap}_{k=1}^{M-1}} Y_k \cap X = \emptyset\\
\end{array}} \\

\frac{\displaystyle \prod_{j=1}^{M-1} m_{\sigma_i(j)}(Y_{\sigma_i(j)})}{\displaystyle m_i(X)+\sum_{j=1}^{M-1} m_{\sigma_i(j)}(Y_{\sigma_i(j)})},
\end{array}
\end{eqnarray}
where:
\begin{eqnarray}
\left\{
\begin{array}{l}
\sigma_i(j)=j, ~~\mbox{if}~~j<i,\\
\sigma_i(j)=j+1, ~~\mbox{if}~~j\geq i,\\
\end{array}
\right.
\end{eqnarray}
$m_i(X)+\displaystyle \sum_{j=1}^{M-1} m_{\sigma_i(j)}(Y_{\sigma_i(j)}) \neq 0$, $m_c$ is the conjunctive consensus rule given by the equation (\ref{conjunctive}). This rule allows a proportional conflict redistribution on the subsets from where the conflict comes and is equivalent for two experts to the rule given in \cite{Smarandache05}. This rule will be illustrated on simple examples in the next section.

These rules are compared in \cite{Osswald06}.

\subsection{Decision level}

The decision is a difficult task. No measures are able to provide the best decision in all the cases. Generally, we consider the maximum of one of the three functions: credibility, plausibility, and pignistic probability. 

The pignistic probability, introduced by \cite{Smets90b}, is here given for all $X \in 2^\Theta$, with $X \neq \emptyset$ by:
\begin{eqnarray}
\betP(X)=\sum_{Y \in 2^\Theta, Y \neq \emptyset} \frac{|X \cap Y|}{|Y|} \frac{m(Y)}{1-m(\emptyset)}.
\end{eqnarray}

If the credibility function provides a pessimist decision, the plausibility function is often too optimist. The pignistic probability is usually taken as a compromise. 

\section{Experts fusion}
\label{expert_fusion}

In order to fuse the opinions of different experts on a given tile $X$, we have to take into account the certainty of experts and proportion of the two (or more) sediments but not only on one focal element. In this case, the space of discernment $\Theta$ represents the different kind of sediments on sonar images, such as rock, sand, silt, cobble, ripple or shadow (that means no sediment information). The experts give their perception and belief according to their certainty. For instance, the expert can be moderately sure of his choice when he labels one part of the image as belonging to a certain class, and be totally doubtful on another part of the image. Moreover, on a considered tile, more than one sediment can be present. 

Consequently we have to take into account all these aspects of the applications. In order to simplify, we consider only two classes in the following: the rock referred as $A$, and the sand, referred as $B$. The proposed models can be easily extended, but their study is easier to understand with only two classes.

Hence, on certain tiles, $A$ and $B$ can be present for one or more experts. The belief functions have to take into account the certainty given by the experts (referred respectively as $c_A$ and $c_B$, two numbers in $[0,1]$) as well as the proportion of the kind of sediment in the tile $X$ (referred as $p_A$ and $p_B$, also two numbers in $[0,1]$). We have two interpretations of ``the expert believes $A$'': it can mean that the expert thinks that there is $A$ on $X$ and not $B$, or it can mean that the expert thinks that there is $A$ on $X$ and it can also have $B$ but he does not say anything about it. The first interpretation yields that hypotheses $A$ and $B$ are exclusive and with the second they are not exclusive. We only study the first case: $A$ and $B$ are exclusive. But on the tile $X$, the expert can also provide $A$ and $B$, in this case the two propositions ``the expert believes $A$'' and ``the expert believes $A$ and $B$'' are not exclusive.

We propose a model considering only one belief function according to the proportion by:
\begin{eqnarray}
\label{M5}
\begin{array}{l}
  \left\{
  \begin{array}{l}
  m(A)=p_A.c_A, \\
  m(B)=p_B.c_B,\\
  m(A \cup B)=1-(p_A . c_A + p_B . c_B).
  \end{array}
  \right. \\
\end{array}
\end{eqnarray}

For instance, consider two experts providing their opinion on the tile $X$. The first expert says that on tile $X$ there is some rock $A$ with a certainty equal to 0.6. Hence for this first expert we have : $p_A=1$, $p_B=0$, and $c_A=0.6$. The second expert thinks that there are 50\% of rock and 50\% of sand on the considered tile $X$ with a respective certainty of 0.6 and 0.4. Hence for the second expert we have: $p_A=0.5$, $p_B=0.5$, $c_A=0.6$ and $c_B=0.4$. We illustrate all our proposed models with this numerical exemple.

Consequently, we have simply:
\begin{eqnarray*}
  \begin{array}{|c|c|c|c|}
  \hline
   & A & B & A \cup B \\
  \hline
   m_1 & 0.6 & 0  & 0.4 \\
  \hline
   m_2 & 0.3& 0.2 & 0.5 \\
  \hline  
  \end{array}
\end{eqnarray*}

The non-normalized conjunctive rule, the credibility, the plausibility and the pignistic probability are given by:
\begin{eqnarray*}
  \begin{array}{|c|c|c|c|c|}
  \hline
  element  & m_c & \bel & \pl & \betP \\
  \hline
\emptyset & 0.12 & 0  & 0 & - \\
  \hline
A & 0.6&  0.6 & 0.8 & 0.7955 \\
  \hline
B & 0.08 & 0.08& 0.28& 0.2045 \\
  \hline
A\cup B & 0.2 & 0.88  & 0.88 & 1 \\
  \hline
  \end{array}
\end{eqnarray*}
In this case we do not have the possibility to decide on $A\cap B$, because the conflict is on $\emptyset$.

The PCR rule provides:
\begin{eqnarray*}
  \begin{array}{|c|c|c|c|c|}
  \hline
  element  & m_{PCR} & \bel & \pl & \betP \\
  \hline
\emptyset & 0 & 0  & 0 & - \\
  \hline
A & 0.69&  0.69 & 0.89 & 0.79 \\
  \hline
B & 0.11 & 0.11& 0.31& 0.21 \\
  \hline
A\cup B & 0.2 & 1  & 1 & 1 \\
  \hline
  \end{array}
\end{eqnarray*}
where
\begin{eqnarray*}
  \begin{array}{l}
  m_{PCR}(A)=0.60+0.09=0.69, \\
  m_{PCR}(B)=0.08+0.03=0.11. \\
  \end{array}
\end{eqnarray*}
With the PCR rule, the decision will be also $A$.

Of course, we cannot say on this example which rule is the best, and we can apply these two rules in order to construct a reality taking into account the doubts of different experts. This reality can serve to train a classifier and also to evaluate this classifier. We can use many supervised classifiers. In the next section, we propose to introduced a new classifier: a multilayer perceptron based on belief learning, take into account all the reachness of the belief basic assignment.

\section{Multilayer Perceptron based on Belief Learning}
We propose in this section a new belief multilayer perceptron where the difference between the multilayer perceptron relates to the learning based on a belief learning. In \cite{Denoeux00a}, a neural network classifier based on Dempster-Shafer theory is presented. In this work, the neural network consider the bba at each neuron, that is not the case in our approach presented feedforward.

\subsection{A multilayer perceptron}
The neural network classifiers are today the most used supervised classifiers. The multilayer perceptron (MLP) is a feedforward fully connected neural network. 
%
%

\begin{figure}[htb]
\begin{center}
\includegraphics[height=3cm]{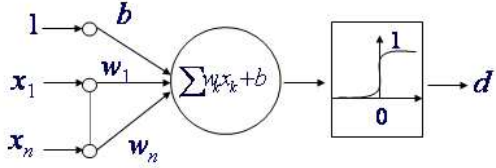}
\end{center}
\caption{Artificial neuron structure.}
\label{neuron}
\end{figure}

The tile $X$ is described by $n$ features $(x_1, ..., x_n)$. Each unit of the network is an artificial neuron called perceptron, with the structure given in figure \ref{neuron}. 

All the neuron outputs of every layer are connected to all the neuron inputs of the next layer weighted by values we have to learn. These weights are first initialized with small random values. In order to learn these values we present to the network the learning vectors and the corresponding desired outputs. The objective of the learning process is to minimize the quadratic error:
\begin{eqnarray}
\label{min_error}
\epsilon=\frac{1}{2}\sum_{i=1}^N (d_i - s_i)^2,
\end{eqnarray}
where $s_i$ are the obtained outputs of the multilayer perceptron and $d_i$ is 1 if the class of $X$ is $C_i$ and 0 elsewhere. As shown on figure \ref{neuron}, we can use the sigmoid function given by:
\begin{eqnarray}
f(x)=\frac{1}{1+e^{-x}}.
\end{eqnarray}
So we obtain the learning algorithm called the back propagation algorithm for the iteration $t+1$:
\begin{eqnarray}
w_{l_1l_2}(t+1)=w_{l_1l_2}(t)+\eta \delta_{l_2}(t) s_{l_1}(t),
\end{eqnarray}
where $w_{l_1l_2}$ is the weight value between the neuron $l_1$ of the first layer and the neuron $l_2$ of the following layer, $\eta$ stands for the learning rate, $s_{l_1}(t)$ is the obtained output of the neuron $l_1$ at the iteration $t$, and $\delta_{l_2}(t)$ is given by:
\begin{eqnarray}
\delta_i(t)=c s_i(t) (1-s_i(t))(d_i-s_i(t)),
\end{eqnarray}
if $l_2=i$ is on the output layer, where the constant $c$ controls the slope of the sigmoid function, and
\begin{eqnarray}
\delta_{l_2}(t)=c s_{l_2}(t) (1-s_{l_2}(t))\sum_l \delta_{l}(t)  w_{ll_2}(t),
\end{eqnarray}
elsewhere.

\subsection{Belief learning}
The use of uncertain and imprecise data for learning have been used in \cite{Denoeux00,Vannoorenberghe02} for decision trees and in \cite{Ambroise01,Vannoorenberghe05} for a credal EM approach. In the previous approach, the learning set ${\cal L}$ is composed of $K$ examples $(X_t,C_t)$, $t=1, ..., K$, where $X_t$ is a tile (a $n$-dimensional  vector given by $n$ features calculated on the tile) and $C_t \in \Theta$ the class of $X_t$. The learning set is also given by the couples $(X_t,d_t)$, with $d_t$ the function equal to 1 if the class of $X$ is $C_t$ and 0 elsewhere. The belief learning is based on the use of a learning set ${\tilde{\cal L}}$ given by:
\begin{equation}
{\tilde{\cal L}}=\{(X_t,m_t^\Theta), t=1,..., K\},
\end{equation}
where $m_t^\Theta$ is the bba defined on $\Theta$. 

In our case, human expert cannot provide with certainty the class of a given tile $X$, and according to the experts, more than one class can be present on the tile $X$. Hence we cannot have the function $d_i$ that is 1 if the class of $X$ is $C_i$ and 0 elsewhere.

The simple idea of the belief learning for the multilayer perceptron is to consider the belief basic assignment in order to minimize the error $\epsilon$ given by the equation (\ref{min_error}). Hence, we obtain $2^{|\Theta |}$ neurons on the output level and we can stay in the credal level.

\subsection{Decision level}

Usually the decision is taken considering the maximum of the values on the output layer. These values are between 0 and 1, but the sum is not 1. We can easily normalize them in order to interpret these values as belief basic assignment. For instance the normalization can be made dividing by the sum of the values of the output layer. Hence, the decision can be conducted by the maximum of the pignistic probability, or with other function such as the credibility or the plausibility. Note that if the output layer is composed only with the singletons, to consider the maximum of the values or the maximum of the pignistic probability is the same.

\section{Illustration}
\label{illustration}

\subsection{Database}
Our database contains 42 sonar images provided by the GESMA (Groupe d'Etudes Sous-Marines de l'Atlantique). These images were obtained with a Klein 5400 lateral sonar with a resolution of 20 to 30 cm in azimuth and 3 cm in range. The sea-bottom depth was between 15 m and 40 m.

\begin{figure}[htb]
\begin{center}
\includegraphics[height=5cm]{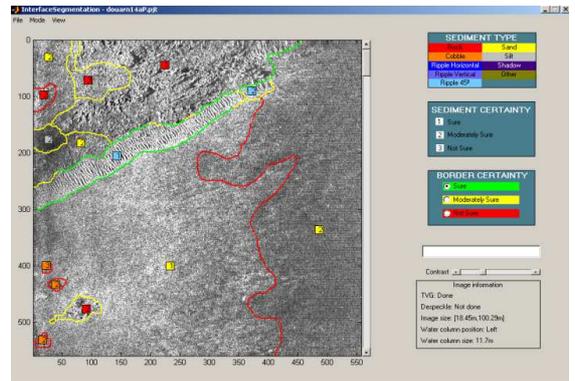}
\end{center}
\caption{Manual Segmentation Interface.}
\label{manual_seg}
\end{figure}

Three experts have manually segmented these images giving the kind of sediment (rock, cobble, sand, silt, ripple (horizontal, vertical or at 45 degrees)), shadow or other (typically ships) parts on images, helped by the manual segmentation interface presented in figure \ref{manual_seg}. All sediments are given with a certainty level (sure, moderately sure or not sure), and the boundary between two sediments is also given with a certainty (sure, moderately sure or not sure). Hence, every pixel of every image is labeled as being either a certain type of sediment or a shadow or other, or a boundary with one of the three certainty levels. We choose the weights: 2/3, 1/2 and 1/3, for respectively the certainty levels: sure, moderately sure and not sure.

\subsection{Experts Fusion}
In order to obtain a kind of reality for learning task, we first fuse the opinion of the three experts following the presented model. We note $A$ for rock, $B$ for sand, $C$ for cobble, $D$ for silt, $E$ for ripple, $F$ for shadow and $G$ for other, hence we have seven classes and $\Theta=\{A,B,C,D,E,F,G\}$. We have applied our model on tiles of size 64$\times$64 pixels given by:
\begin{eqnarray}
\label{GeneralizedM5}
\begin{array}{l}
  \left\{
  \begin{array}{l}
  m(A)=p_{A1}.c_1+p_{A2}.c_2+p_{A3}.c_3\\
  m(B)=p_{B1}.c_1+p_{B2}.c_2+p_{B3}.c_3\\
  m(C)=p_{C1}.c_1+p_{C2}.c_2+p_{C3}.c_3\\
  m(D)=p_{D1}.c_1+p_{D2}.c_2+p_{D3}.c_3\\
  m(E)=p_{E1}.c_1+p_{E2}.c_2+p_{E3}.c_3\\
  m(F)=p_{F1}.c_1+p_{F2}.c_2+p_{F3}.c_3 \\
  m(G)=p_{G1}.c_1+p_{G2}.c_2+p_{G3}.c_3\\
  m(\Theta)=1-(m(A)+m(B)+m(C)\\
  ~~~~~~~~+m(D)+m(E)+m(F)+m(G)),\\
  \end{array}
  \right. \\
\end{array}
\end{eqnarray}
where $c_1$, $c_2$ and $c_3$ are the weights associated to the certitude respectively: ``sure'', ``moderately sure'' and ``not sure'' ({\em e.g.} here:  $c_1=2/3$, $c_2=1/2$ and $c_3=1/3$). Indeed we have to consider the cases when the same kind of sediment (but with different certainties) is present on the same tile. The proportion of each sediment in the tile associated to these weights is noted, for instance for $A$: $p_{A1}$, $p_{A2}$ and $p_{A3}$. 

In order to provide a reality for the learning, the experts can be fuse by the non-normalized conjunctive rule or the generalized PCR as we see before, and the decision can be taken on the maximum of the pignistic probability. The total conflict between the three experts is 0.2432. This conflict comes essentially from the difference of opinion of the experts and not from the tiles with more than one sediment. Indeed, we have a weak {\it auto-conflict} (conflict coming from the combination of the same expert three times). The values of the auto-conflict for the three experts are: 0.0841, 0.0840, and 0.0746. We note a difference of decision between the three combination rules giving by the equations (\ref{GeneDSmTcombination}) for the generalized PCR, and (\ref{conjunctive}) for the conjunctive rule. The proportion of tiles with a different decision is 1.01\% between the generalized PCR and the conjunctive rule. However, we cannot evaluate on this database which combination rule is the best.

\subsection{Results}
In order to classify the tiles of size 64$\times$64 pixels, we first have to extract texture parameters from each tile. Here, we choose the co-occurrence matrices approach \cite{Martin05}. The co-occurrence matrices are calculated by numbering the occurrences of identical gray level of two pixels. Four directions are considered: 0, 45, 90 and 135 degrees. Concerning these four directions, six parameters given by \cite{Haralick79} are calculated: homogeneity, contrast estimation, entropy estimation, the correlation, the directivity, and the uniformity. This classical approach yields 24 parameters. The problem for co-occurrence matrices is the non-invariance in translation. Typically, this problem can appear in a ripple texture characterization. More features extraction approaches can be used such as the run-lengths matrix, the wavelet transform and the Gabor filters \cite{Martin05}.

Hence, each tile is represented by the 24 parameters, and we can try to classify the tiles by the multilayer perceptron and the belief multilayer perceptron. So, the input layer contains 24 neurons, and the output layer contains 7 neurons (one for each class). For the belief multilayer perceptron, the mass calculated by the fusion of the three experts according to the model given in (\ref{GeneralizedM5}) allows the learning. We test the both combination given by the conjunctive non-normalized rule (\ref{conjunctive}) and the PCR rule (\ref{GeneDSmTcombination}). The mass model gives focal element only on the singleton and the ignorance $\Theta$. In order to learn only on the singletons, we consider only the bba given on the singletons, and we renormalize them in order to obtain one for the singleton given the maximum belief; the output values are not bba in all the case.

Hence, the output layer of the belief multilayer perceptron is composed only by seven neurons (one for each class). Of course, it could be more interesting to keep $2^\Theta=128$ neurons on the last layer in order to stay in the credal level and keep the power of this classifier. However, this is possible only if enough data are available for the learning.

In order to take a decision on bba with the maximum of the pignistic probability, we annul the minimum value of the output layer then we normalize by the sum of the values. Here it is similar to decide on the maximum of the values of the output layer, but it is not the same in all the cases as shown afterwards.

On the 42 sonar images, we have 9266 tiles of size 64$\times$64 pixels. Our database has been randomly divided into two parts. The first one (2/3 of the database) is used for the multilayer perceptron and the belief multilayer perceptron learnings, and the second one for tests. We repeat this random division 30 times in order to achieve a good estimation of the classification rate, and we analyze the mean percentage of good classification rates defined as the number of good classified small-images dived by the total of small-images.

With the non-normalized conjunctive rule, we obtain 64.49\% of good-classification rates (with a confidence interval of [64.07;64.91]) for the classic multilayer perceptron and 65.10\% of good-classification rates (with a confidence interval of [64.72;65.48]) for the belief multilayer perceptron. If the reality is obtained by the generalized PCR, we have 64.96\% of good-classification rates (with a confidence interval of [64.44;65.25]) for the classic multilayer perceptron and 64.84\% of good-classification rates (with a confidence interval of [64.55;65.39]) for the belief multilayer perceptron. 

The evaluation is made on an unknown reality and so we can not say that the experts fusion given by the non-normalized conjunctive rule is better than the experts fusion obtained by the generalized PCR rule. In the case of the non-normalized conjunctive rule, the belief multilayer perceptron gives significantly better good-classification rates than the multilayer perceptron. In the case of the generalized PCR rule, the results are not significantly different. However, if we repeat the random division 1000 times, we obtain 65.043\% of good-classification rates (with a confidence interval of [64.97;65.11]) for the classic multilayer perceptron and 65.125\% of good-classification rates (with a confidence interval of [65.06;65.19]) for the belief multilayer perceptron, with the reality is obtained by the generalized PCR. These results show that, here also, the belief multilayer perceptron improves significantly the classification rates. 

Another interest of the belief multilayer perceptron comes from the decision step. For instance, the cobble can be seen like a doubt between the rock and the sand according to the size of the tile. Hence, the class $C$ can be rewriten as $A\cup B$. The learning will be the same that previously, but the decision by the maximum of the probability pignistic will provide another result. We cannot compare these results with the classic multilayer perceptron, because the decision step is taken by the maximum of the values of the outputs layer. Another example can be done if we consider the class shadow like the absence of information, {\em e.g.} we can associate this class to the ignorance and rewrite the model by:
\begin{eqnarray}
\label{GeneralizedM5_2}
\begin{array}{l}
  \left\{
  \begin{array}{l}
  m(A)=p_{A1}.c_1+p_{A2}.c_2+p_{A3}.c_3\\
  m(B)=p_{B1}.c_1+p_{B2}.c_2+p_{B3}.c_3\\
  m(C)=p_{C1}.c_1+p_{C2}.c_2+p_{C3}.c_3\\
  m(D)=p_{D1}.c_1+p_{D2}.c_2+p_{D3}.c_3\\
  m(E)=p_{E1}.c_1+p_{E2}.c_2+p_{E3}.c_3\\
  m(G)=p_{G1}.c_1+p_{G2}.c_2+p_{G3}.c_3\\
  m(\Theta)=1-(m(A)+m(B)+m(C)\\
  ~~~~~~~~+m(D)+m(E)+m(G)).\\
  \end{array}
  \right. \\
\end{array}
\end{eqnarray}
The ignorance $\Theta$ can be learned and so be represented by a neuron on the output layer. Here also, the decision by the maximum of the probability pignistic provides another result, we cannot compare with the classic multilayer perceptron.

\section{Conclusions}
We have proposed in this paper two different fusion approaches in sonar images processing. The first novelty is the experts fusion model, we can apply in many image processing problems. Indeed, if some images represent uncertain environments, the reality is unknown and we must compose and propose a reality ({\em e.g.} in order to train a classifier) from the experts opinions. In this kind of environments, experts cannot say with certainty what is exactly on the images and we have to take into account the doubt of the experts in order to describe the images. The second novelty is the multilayer perceptron with a belief learning improves significantly the classic multilayer perceptron. It could be more interesting to keep $2^{|\Theta |}$ neurons on the last layer in order to stay in the credal level and keep the power of this classifier. Hence, this classifier can provide a belief on every subset of $2^\Theta$, and the decision can be made on this space. The evaluation of this classifier must be made on more data sets especially on databases where the real classes are known and with data giving in terms of belief. The problem of image classification evaluation is very hard to solve in uncertain environment \cite{Martin06b}.

\end{document}